\documentclass[10pt,twocolumn,letterpaper]{article}

\usepackage{cvpr}
\usepackage{times}
\usepackage{epsfig}
\usepackage{graphicx}
\usepackage{amsmath}
\usepackage{amssymb}

\usepackage[english]{babel}
\usepackage[numbers]{natbib}
\usepackage[super]{nth}
\usepackage{url}

\cvprfinalcopy 

\def\cvprPaperID{1325} 

\ifcvprfinal\fi
\begin{document}

\title{Computing the Stereo Matching Cost with a Convolutional Neural Network}

\author{Jure \v{Z}bontar \\
University of Ljubljana\\
{\tt\small jure.zbontar@fri.uni-lj.si}
\and
Yann LeCun\\
New York University\\
{\tt\small yann@cs.nyu.edu}
}
\maketitle

\begin{abstract}
We present a method for extracting depth information from a rectified image
pair. We train a convolutional neural network to predict how well two image
patches match and use it to compute the stereo matching cost. The cost is
refined by cross-based cost aggregation and semiglobal matching, followed by a
left-right consistency check to eliminate errors in the occluded regions. Our
stereo method achieves an error rate of 2.61\,\% on the KITTI stereo dataset
and is currently (August 2014) the top performing method on this dataset.

\end{abstract}

\section{Introduction} 
Consider the following problem: given two images taken from cameras at
different horizontal positions, the goal is to compute the disparity $d$ for
each pixel in the left image. Disparity refers to the difference in horizontal
location of an object in the left and right image---an object at position $(x,
y)$ in the left image will appear at position $(x - d, y)$ in the right image.
Knowing the disparity $d$ of an object, we can compute its depth $z$ (\ie the
distance from the object to the camera) by using the following relation: 
\begin{equation}
z = \frac{fB}{d},
\end{equation}
where $f$ is the focal length of the camera and $B$ is the
distance between the camera centers.

The described problem is a subproblem of stereo reconstruction, where the goal
is to extract 3D shape from one or more images.  According to the taxonomy of
\citet{scharstein2002taxonomy}, a typical stereo algorithm consists of four
steps: (1) matching cost computation, (2) cost aggregation, (3) optimization,
and (4) disparity refinement.  Following \citet{hirschmuller2009evaluation}, we
refer to steps (1) and (2) as computing the matching cost and steps (3) and (4)
as the stereo method.

\begin{figure*}[ht]
\begin{center}
\includegraphics{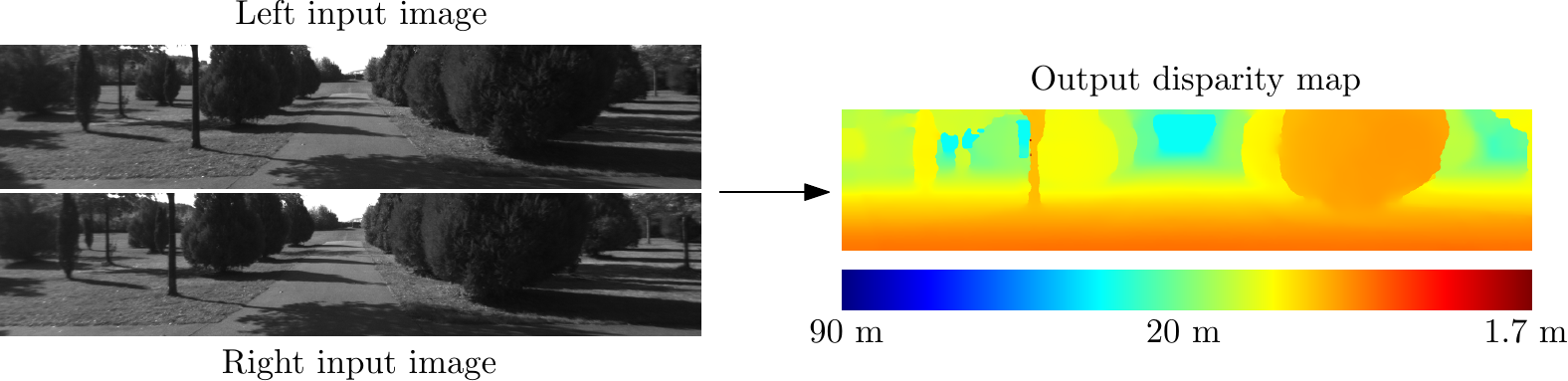}
\end{center}
\caption{The input is a pair of images from the left and right camera. The two
input images differ mostly in horizontal locations of objects. Note that
objects closer to the camera have larger disparities than objects farther away.
The output is a dense disparity map shown on the right, with warmer colors
representing larger values of disparity (and smaller values of depth).}
\label{fig:input_output}

\end{figure*}

We propose training a convolutional neural network~\cite{lecun1998gradient} on
pairs of small image patches where the true disparity is known (\eg obtained by
LIDAR). The output of the network is used to initialize the matching cost
between a pair of patches.  Matching costs are combined between neighboring
pixels with similar image intensities using cross-based cost aggregation.
Smoothness constraints are enforced by semiglobal matching and a left-right
consistency check is used to detect and eliminate errors in occluded regions.
We perform subpixel enhancement and apply a median filter and a bilateral
filter to obtain the final disparity map.  Figure~\ref{fig:input_output}
depicts the inputs to and the output from our method. 
The two contributions of this paper are:

\begin{itemize}
\item We describe how a convolutional neural network can be used to compute
the stereo matching cost.

\item We achieve an error rate of 2.61\,\% on the KITTI stereo dataset,
improving on the previous best result of 2.83\,\%.

\end{itemize}

\section{Related work}

Before the introduction of large stereo
datasets~\cite{Geiger2013IJRR,scharstein2007learning}, relatively few stereo
algorithms used ground-truth information to learn parameters of their models;
in this section, we review the ones that did. For a general overview of stereo
algorithms see~\cite{scharstein2002taxonomy}.

\citet{kong2004method} used sum of squared distances to compute an initial
matching cost. They trained a model to predict the probability distribution over
three classes: the initial disparity is correct, the initial disparity is
incorrect due to fattening of a foreground object, and the initial disparity is
incorrect due to other reasons. The predicted probabilities were used to adjust
the initial matching cost. \citet{kong2006stereo} later extend their work by
combining predictions obtained by computing normalized cross-correlation over
different window sizes and centers.
\citet{peris2012towards} initialized the matching cost with
AD-Census~\cite{mei2011building} and used multiclass linear discriminant
analysis to learn a mapping from the computed matching cost to the final
disparity.

Ground-truth data was also used to learn parameters of graphical models.
\citet{zhang2007estimating} used an alternative optimization algorithm to estimate
optimal values of Markov random field hyperparameters.
\citet{scharstein2007learning} constructed a new dataset of 30 stereo pairs and
used it to learn parameters of a conditional random field.
\citet{li2008learning} presented a conditional random field model with a
non-parametric cost function and used a structured support vector machine to
learn the model parameters.

Recent work~\cite{haeusler2013ensemble,spyropoulos2014learning} focused on
estimating the confidence of the computed matching cost.
\citet{haeusler2013ensemble} used a
random forest classifier to combine several confidence measures. Similarly,
\citet{spyropoulos2014learning} trained a random forest classifier to predict
the confidence of the matching cost and used the predictions as soft constraints
in a Markov random field to decrease the error of the stereo method.

\section{Computing the matching cost}

A typical stereo algorithm begins by computing a matching cost $C(\textbf{p},
d)$ at each position $\textbf{p}$ for all disparities $d$ under consideration.
A simple example is the sum of absolute differences:
\begin{equation} C_{\text{AD}}(\textbf{p}, d) = \sum_{\textbf{q} \in
\mathcal{N}_{\textbf{p}}} |I^L(\textbf{q}) - I^R(\textbf{qd})|,
\label{eqn:C_ad}
\end{equation}
where $I^L(\textbf{p})$ and $I^R(\textbf{p})$ are image intensities at
position $\textbf{p}$ of the left and right image and
$\mathcal{N}_{\textbf{p}}$ is the set of locations within a fixed rectangular
window centered at $\textbf{p}$. We use bold lowercase letters ($\textbf{p}$,
$\textbf{q},$ and $\textbf{r}$) to denote pairs of real numbers. Appending a
lowercase $\textbf{d}$ has the following meaning: if $\textbf{p} = (x, y)$ then
$\textbf{pd} = (x - d, y)$.

Equation~\eqref{eqn:C_ad} can be interpreted as measuring the cost associated
with matching a patch from the left image, centered at position $\textbf{p}$,
with a patch from the right image, centered at position $\textbf{pd}$. Since
examples of good and bad matches can be obtained from publicly available
datasets, \eg KITTI~\cite{Geiger2013IJRR} and
Middlebury~\cite{scharstein2002taxonomy}, we can attempt to solve the matching
problem by a supervised learning approach.  Inspired by the successful
applications of convolutional neural networks to vision
problems~\cite{krizhevsky2012imagenet}, we used them to evaluate how well two
small image patches match.

\subsection{Creating the dataset}
A training example comprises two patches, one from the left and one from the
right image:
\begin{equation}
<\mathcal{P}_{9 \times 9}^L(\textbf{p}), \mathcal{P}_{9 \times 9}^R
(\textbf{q})>,
\end{equation}
where $\mathcal{P}_{9 \times 9}^L(\textbf{p})$ denotes a $9 \times 9$ patch
from the left image, centered at $\textbf{p} = (x, y)$.  For each location
where the true disparity $d$ is known, we extract one negative and one positive
example.  A negative example is obtained by setting the center of the right
patch $\textbf{q}$ to 
\begin{equation}
\textbf{q} = (x - d + o_{\text{neg}}, y),
\end{equation}
where $o_{\text{neg}}$ is an offset corrupting the match, chosen randomly from the
set $\{-N_{\text{hi}}, \ldots, -N_{\text{lo}}, N_{\text{lo}}, \ldots,
N_{\text{hi}}\}$. Similarly, a positive example is derived by setting
\begin{equation}
\textbf{q} = (x - d + o_{\text{pos}}, y),
\end{equation}
where $o_{\text{pos}}$ is chosen
randomly from the set $\{-P_{\text{hi}}, \ldots, P_{\text{hi}}\}$. The reason
for including $o_{\text{pos}}$, instead of setting it to zero, has to do with
the stereo method used later on.  In particular, we found that cross-based cost
aggregation performs better when the network assigns low matching costs to good
matches as well as near matches.  $N_{\text{lo}}$, $N_{\text{hi}}$,
$P_{\text{hi}}$, and the size of the image patches $n$ are hyperparameters of
the method.

\subsection{Network architecture}

\begin{figure}[t]
\begin{center}
\includegraphics[width=0.38\textwidth]{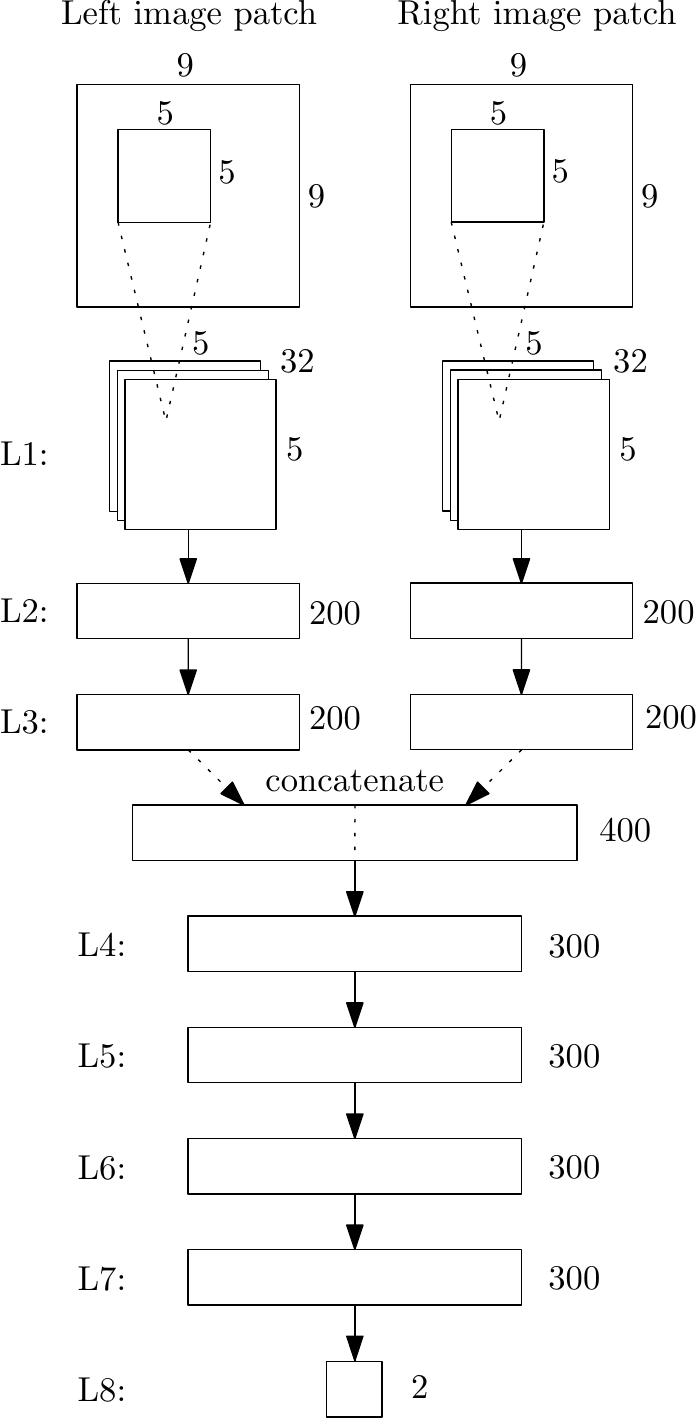}
\end{center}
\caption{The architecture of our convolutional neural network.}
\label{fig:architecture}
\end{figure}

The architecture we used is depicted in Figure \ref{fig:architecture}.  The
network consists of eight layers, $L1$ through $L8$. The first layer is
convolutional, while all other layers are fully-connected. The inputs to the
network are two $9 \times 9$ gray image patches. The first convolutional layer
consists of 32 kernels of size $5 \times 5 \times 1$. Layers $L2$ and $L3$ are
fully-connected with 200 neurons each. After $L3$ the two 200 dimensional
vectors are concatenated into a 400 dimensional vector and passed through four
fully-connected layers, $L4$ through $L7$, with 300 neurons each. The final
layer, $L8$, projects the output to two real numbers that are fed through a
softmax function, producing a distribution over the two classes (good match and
bad match). The weights in $L1$, $L2$, and $L3$ of the networks for the left
and right image patch are tied.  Rectified linear units follow each layer,
except $L8$.  We did not use pooling in our architecture.  The network contains
almost 600~thousand parameters. The architecture is appropriate for gray
images, but can easily be extended to handle RGB images by learning $5 \times 5
\times 3$, instead of $5 \times 5 \times 1$ filters in $L1$. The best
hyperparameters of the network (such as the number of layers, the number of neurons
in each layer, and the size of input patches) will differ from one dataset to
another.  We chose this architecture because it performed well on the KITTI
stereo dataset.

\subsection{Matching cost}

The matching cost $C_{\text{CNN}}(\textbf{p}, d)$ is computed directly from the
output of the network:
\begin{equation}
C_{\text{CNN}}(\textbf{p}, d) = f_{\text{neg}}(<\mathcal{P}_{9 \times
9}^L(\textbf{p}), \mathcal{P}_{9 \times 9}^R(\textbf{pd})>),
\end{equation}
where $f_{\text{neg}}(<\mathcal{P}^L, \mathcal{P}^R>)$ is the output of the
network for the negative class when run on input patches $\mathcal{P}^L$ and
$\mathcal{P}^R$. 

Naively, we would have to perform the forward pass for each
image location $\textbf{p}$ and each disparity $d$ under consideration. The
following three implementation details kept the runtime manageable:

\begin{enumerate}
\item The output of layers $L1$, $L2$, and $L3$ need to be computed only once
per location $\textbf{p}$ and need not be recomputed for every disparity $d$.

\item The output of $L3$ can be computed for all locations in a single forward
pass by feeding the network full-resolution images, instead of $9 \times 9$
image patches. To achieve this, we apply layers $L2$ and $L3$
convolutionally---layer $L2$ with filters of size $5 \times 5 \times 32$ and
layer $L3$ with filters of size $1 \times 1 \times 200$, both outputting 200
feature maps.

\item Similarly, $L4$ through $L8$ can be replaced with convolutional filters
of size $1 \times 1$ in order to compute the output of all locations in a
single forward pass. Unfortunately, we still have to perform the forward pass
for each disparity under consideration.

\end{enumerate}

\section{Stereo method}
\label{sec:stereo_method}

In order to meaningfully evaluate the matching cost, we need to pair it with a
stereo method. The stereo method we used was influenced by
\citet{mei2011building}.

\subsection{Cross-based cost aggregation}

Information from neighboring pixels can be combined by averaging the
matching cost over a fixed window.  This approach fails near depth
discontinuities where the assumption of constant depth within a window is
violated. We might prefer a method that adaptively selects the neighborhood for
each pixel so that support is collected only from pixels with similar
disparities. In cross-based cost aggregation \cite{zhang2009cross} we build a
local neighborhood around each location comprising pixels with similar
image intensity values.

\begin{figure}[ht]
\begin{center}
\includegraphics[width=0.4\textwidth]{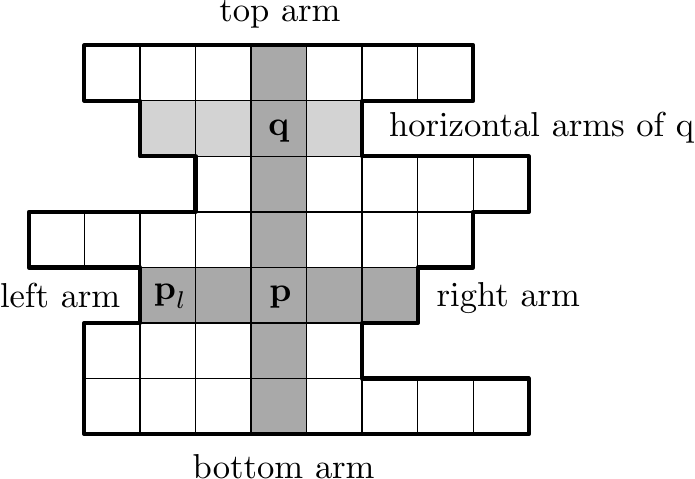}
\end{center}
\caption{The support region for position $\textbf{p}$, is the union of
horizontal arms of all positions $\textbf{q}$ on $\textbf{p}$'s vertical arm.}
\label{fig:cross}
\end{figure}

Cross-based cost aggregation begins by constructing an upright cross at each
position. The left arm $\textbf{p}_l$ at position $\textbf{p}$ extends left as
long as the following two conditions hold:

\begin{itemize}

\item $|I(\textbf{p}) - I(\textbf{p}_l)| < \tau$. The absolute difference in
image intensities at positions $\textbf{p}$ and $\textbf{p}_l$ is smaller than
$\tau$.  

\item $\|\textbf{p} - \textbf{p}_l\| < \eta$. The horizontal distance (or
vertical distance, in case of top and bottom arms) between $\textbf{p}$ and
$\textbf{p}_l$ is less than $\eta$.  

\end{itemize}
The right, bottom, and top arms are constructed analogously. Once the four arms
are known, we can define the support region $U(\textbf{p})$ as the union of
horizontal arms of all positions $\textbf{q}$ laying on $\textbf{p}$'s
vertical arm (see Figure \ref{fig:cross}).
\citet{zhang2009cross} suggest that aggregation should consider the support
regions of both images in a stereo pair. Let $U^L$ and $U^R$ denote the support
regions in the left and right image. We define the combined support region
$U_d$ as
\begin{equation}
U_d(\textbf{p}) = \{\textbf{q} | \textbf{q} \in U^L(\textbf{p}), \textbf{qd}
\in U^R(\textbf{pd})\}.
\end{equation}
The matching cost is averaged over the combined support region:

\begin{align}
C^0_{\text{CBCA}}(\textbf{p}, d) &= C_{\text{CNN}}(\textbf{p}, d), \\
C^i_{\text{CBCA}}(\textbf{p}, d) &= \frac{1}{|U_d(\textbf{p})|}
\sum_{\textbf{q} \in U_d(\textbf{p})} C^{i-1}_{\text{CBCA}}(\textbf{q}, d),
\end{align}

where $i$ is the iteration number. We repeat the averaging four times; the output
of cross-based cost aggregation is $C^4_{\text{CBCA}}$.

\subsection{Semiglobal matching}

We refine the matching cost by enforcing smoothness constraints on the
disparity image. Following~\citet{hirschmuller2008stereo}, we define an energy
function $E(D)$ that depends on the disparity image $D$:
\begin{multline} E(D) = \sum_{\textbf{p}} \biggl( C^4_{\text{CBCA}}(\textbf{p},
D(\textbf{p})) \\ + \sum_{\textbf{q} \in \mathcal{N}_{\textbf{p}}} P_1 \times
1\{|D(\textbf{p}) - D(\textbf{q})| = 1\} \\ + \sum_{\textbf{q} \in
\mathcal{N}_{\textbf{p}}} P_2 \times 1\{|D(\textbf{p}) - D(\textbf{q})| > 1\}
\biggr), 
\end{multline}
where $1\{\cdot\}$ denotes the indicator function. The first term penalizes
disparities $D(\textbf{p})$ with high matching costs. The second term adds a
penalty $P_1$ when the disparity of neighboring pixels differ by one.  The
third term adds a larger penalty $P_2$ when the neighboring disparities differ
by more than one.  Rather than minimizing $E(D)$ in 2D, we perform the
minimization in a single direction with dynamic programming.  This solution
introduces unwanted streaking effects, since there is no incentive to make the
disparity image smooth in the directions we are not optimizing over. In
semiglobal matching we minimize the energy $E(D)$ in many directions and
average to obtain the final result. Although~\citet{hirschmuller2008stereo}
suggests choosing sixteen direction, we only optimized along the two horizontal
and the two vertical directions; adding the diagonal directions did not improve
the accuracy of our system. 

To minimize $E(D)$ in direction $\textbf{r}$, we define a matching cost
$C_{\textbf{r}}(\textbf{p}, d)$ with the following recurrence relation:
\begin{multline} C_{\textbf{r}}(\textbf{p}, d) = C^4_{\text{CBCA}}(\textbf{p},
d) - \min_k C_r(\textbf{p} - \textbf{r}, k)  \\ + \min\biggl\{ C_r(\textbf{p} -
\textbf{r}, d), C_r(\textbf{p} - \textbf{r}, d - 1) + P_1, \\ C_r(\textbf{p} -
\textbf{r}, d + 1) + P_1, \min_k C_{\textbf{r}}(\textbf{p} - \textbf{r}, k) +
P_2 \biggr\}.  \end{multline}
The second term is included to prevent values of $C_\textbf{r}(\textbf{p}, d)$
from growing too large and does not affect the optimal disparity map.  The
parameters $P_1$ and $P_2$ are set according to the image gradient so that
jumps in disparity coincide with edges in the image. Let $D_1 =
|I^L(\textbf{p}) - I^L(\textbf{p} - \textbf{r})|$ and $D_2 = |I^R(\textbf{pd})
- I^R(\textbf{pd} - \textbf{r})|$.  We set $P_1$ and $P_2$ according to the
following rules:
\[ \begin{array}{lll} P_1 = \Pi_1, &P_2 = \Pi_2 & \text{if $D_1 <
\tau_{\text{SO}}, D_2 < \tau_{\text{SO}}$}, \\ P_1 = \Pi_1 / 4, &P_2 = \Pi_2 /
4 & \text{if $D_1 \geq \tau_{\text{SO}}, D_2 < \tau_{\text{SO}}$}, \\ P_1 =
\Pi_1 / 4, &P_2 = \Pi_2 / 4 & \text{if $D_1 < \tau_{\text{SO}}, D_2 \geq
\tau_{\text{SO}}$}, \\ P_1 = \Pi_1 / 10, &P_2 = \Pi_2 / 10 & \text{if $D_1 \geq
\tau_{\text{SO}}, D_2 \geq \tau_{\text{SO}}$}; \\ \end{array} \]
where $\Pi_1$, $\Pi_2$, and $\tau_\text{SO}$ are hyperparameters.  The value of
$P_1$ is halved when minimizing in the vertical directions. The final cost
$C_\text{SGM}(\textbf{p}, d)$ is computed by taking the average across all four
directions:
\begin{equation} C_\text{SGM}(\textbf{p}, d) = \frac{1}{4} \sum_{\textbf{r}}
C_{\textbf{r}}(\textbf{p}, d).  \end{equation}
After semiglobal matching we repeat cross-based cost aggregation, as described
in the previous section.

\subsection{Computing the disparity image}

The disparity image $D$ is computed by the winner-take-all
strategy, \ie by finding the disparity $d$ that minimizes
$C(\textbf{p}, d)$,
\begin{equation}
D(\textbf{p}) = \arg\!\min_d C(\textbf{p}, d).
\end{equation}

\subsubsection{Interpolation}
Let $D^L$ denote the disparity map obtained by treating the left image as the
reference image---this was the case so far, \ie $D^L(\textbf{p}) =
D(\textbf{p})$---and let $D^R$ denote the disparity map obtained by treating the
right image as the reference image.  Both $D^L$ and $D^R$ contain errors in
occluded regions. We attempt to detect these errors by performing a left-right
consistency check. We label each position $\textbf{p}$ as either
\[ \begin{array}{ll} 
correct & \text{if $|d - D^R(\textbf{pd})| \leq 1$ for $d = D^L(\textbf{p})$}, \\ 
mismatch & \text{if $|d - D^R(\textbf{pd})| \leq 1$ for any other $d$}, \\ 
occlusion & \text{otherwise}. \\ 
\end{array} \]
For positions marked as \emph{occlusion}, we want the new disparity value to
come from the background. We interpolate by moving left until we find a
position labeled \emph{correct} and use its value.  For positions marked as
\emph{mismatch}, we find the nearest \emph{correct} pixels in 16 different
directions and use the median of their disparities for interpolation.  We refer
to the interpolated disparity map as $D_{\text{INT}}$.

\subsubsection{Subpixel enhancement}

Subpixel enhancement provides an easy way to increase the resolution of a
stereo algorithm.  We fit a quadratic curve through the neighboring costs to
obtain a new disparity image:
\begin{equation}
D_{\text{SE}}(\textbf{p}) = d - \frac {C_+ - C_-} {2 (C_+ - 2 C + C_-)},
\end{equation}
where
$d = D_{\text{INT}}(\textbf{p})$,
$C_- = C_{\text{SGM}}(\textbf{p}, d - 1)$,
$C   = C_{\text{SGM}}(\textbf{p}, d    )$, and
$C_+ = C_{\text{SGM}}(\textbf{p}, d + 1)$.

\subsubsection{Refinement}

The size of the disparity image $D_{\text{SE}}$ is smaller than the size of the
original image, due to the bordering effects of convolution. The disparity
image is enlarged to match the size of the input by copying the disparities of
the border pixels.  We proceed by applying a $5 \times 5$ median filter and the
following bilateral filter:
\begin{multline} 
D_{\text{BF}}(\textbf{p}) = \frac{1}{W(\textbf{p})}
\sum_{\textbf{q} \in \mathcal{N}_\textbf{p}} D_{\text{SE}}(\textbf{q}) \cdot
g(\|\textbf{p} - \textbf{q}\|) \\ \cdot 1\{|I^L(\textbf{p}) - I^L(\textbf{q})|
< \tau_{\text{BF}}\},
\end{multline}
where $g(x)$ is the probability density function of a zero mean normal
distribution with standard deviation $\sigma$ and $W(\textbf{p})$ is the
normalizing constant:
\begin{equation}
W(\textbf{p}) = \sum_{\textbf{q} \in \mathcal{N}_\textbf{p}} g(\|\textbf{p} -
\textbf{q}\|) \cdot 1\{|I^L(\textbf{p}) - I^L(\textbf{q})| <
\tau_{\text{BF}}\}.
\end{equation}
$\tau_{\text{BF}}$ and $\sigma$ are hyperparameters. $D_{\text{BF}}$ is the
final output of our stereo method.

\section{Experimental results}

We evaluate our method on the KITTI stereo dataset, because of its large
training set size required to learn the weights of the convolutional neural
network.

\subsection{KITTI stereo dataset}

The KITTI stereo dataset \cite{Geiger2013IJRR} is a collection of gray image
pairs taken from two video cameras mounted on the roof of a car, roughly 54
centimeters apart. The images are recorded while driving in and around the city
of Karlsruhe, in sunny and cloudy weather, at daytime. The dataset comprises 194
training and 195 test image pairs at resolution $1240 \times 376$. Each image
pair is rectified, \ie transformed in such a way that an object appears on the
same vertical position in both images. A rotating laser scanner, mounted behind
the left camera, provides ground truth depth. The true disparities for the test
set are withheld and an online
leaderboard\footnote{\url{http://www.cvlibs.net/datasets/kitti/eval\_stereo\_flow.php?benchmark=stereo}}
is provided where researchers can evaluate their method on the test set.
Submissions are allowed only once every three days. The goal of the KITTI
stereo dataset is to predict the disparity for each pixel on the left image.
Error is measured by the percentage of pixels where the true disparity and the
predicted disparity differ by more than three pixels. Translated into depth,
this means that, for example, the error tolerance is $\pm 3$ centimeters for
objects 2 meters from the camera and $\pm 80$ centimeters for objects 10 meters
from the camera.

\subsection{Details of learning}

We train the network using stochastic gradient descent to minimize the
cross-entropy loss. The batch size was set to 128. We trained for 16 epochs
with the learning rate initially set to 0.01 and decreased by a factor of 10 on
the \nth{12} and \nth{15} iteration. We shuffle the training examples prior to
learning. From the 194 training image pairs we extracted 45 million examples.
Half belonging to the positive class; half to the negative class. We
preprocessed each image by subtracting the mean and dividing by
the standard deviation of its pixel intensity values. The stereo method is
implemented in CUDA, while the network training is done with the Torch7
environment~\cite{collobert2011torch7}.  The hyperparameters of the stereo
method were:
\begin{align*} 
N_{\text{lo}} & =4, & \eta & =4, & \Pi_1 & =1, & \sigma & =5.656, \\
N_{\text{hi}} & =8, & \tau & =0.0442, & \Pi_2 & =32, & \tau_{\text{BF}} & =5, \\
P_{\text{hi}} & =1, & & & \tau_{\text{SO}} &= 0.0625. 
\end{align*} 

\subsection{Results}

Our method achieves an error rate of $2.61\,\%$ on the KITTI stereo test set and is
currently ranked first on the online leaderboard. Table~\ref{tab:kitti_error}
compares the error rates of the best performing stereo algorithms on this dataset.

\begin{table}[ht]
\begin{center}
\begin{tabular}{clll}
\textbf{Rank} & \textbf{Method} & & \textbf{Error} \\\hline
1 & MC-CNN & This paper & 2.61\,\% \\
2 & SPS-StFl & \citet{yamaguchi2014efficient} & 2.83\,\% \\
3 & VC-SF & \citet{vogel2014view} & 3.05\,\% \\
4 & CoP & Anonymous submission & 3.30\,\% \\
5 & SPS-St & \citet{yamaguchi2014efficient} & 3.39\,\% \\
6 & PCBP-SS & \citet{yamaguchi2013robust} & 3.40\,\% \\
7 & DDS-SS & Anonymous submission & 3.83\,\% \\
8 & StereoSLIC & \citet{yamaguchi2013robust} & 3.92\,\% \\
9 & PR-Sf+E & \citet{vogel2013piecewise} & 4.02\,\% \\
10 & PCBP & \citet{yamaguchi2012continuous} & 4.04\,\% \\
\end{tabular}
\end{center}
\caption{The KITTI stereo leaderboard as it stands in November 2014.}
\label{tab:kitti_error}
\end{table}

\begin{figure*}[ht]
\begin{center}
\setlength{\tabcolsep}{1pt}
\renewcommand{\arraystretch}{0.5}
\begin{tabular}{ll}
\includegraphics[scale=0.20]{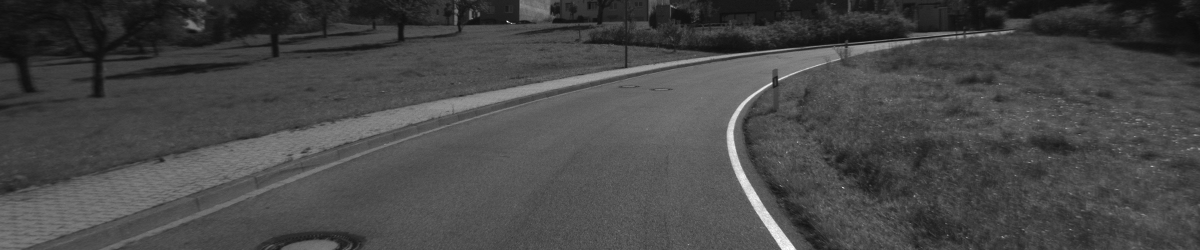} & \includegraphics[scale=0.20]{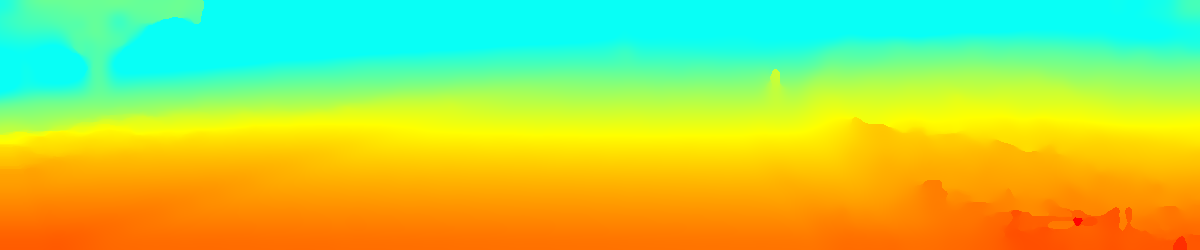} \\
\includegraphics[scale=0.20]{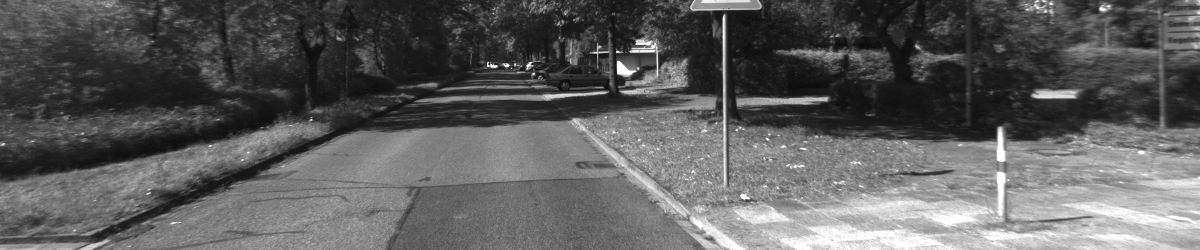} & \includegraphics[scale=0.20]{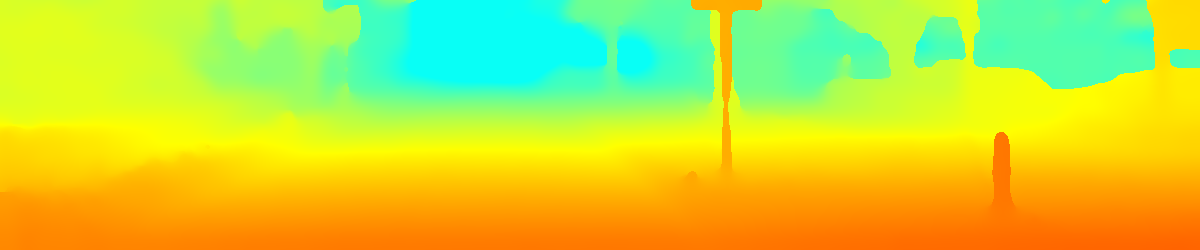} \\
\includegraphics[scale=0.20]{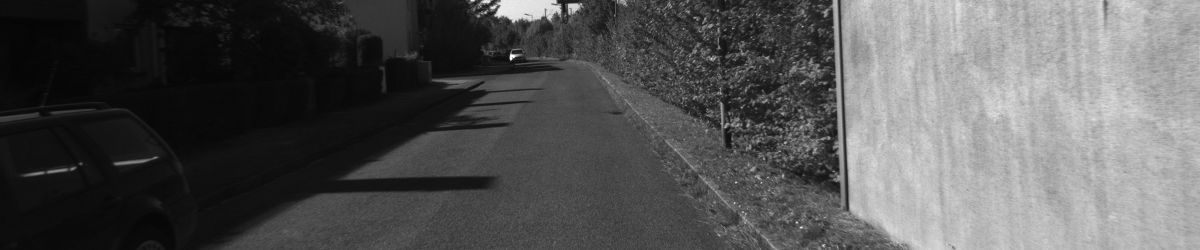} & \includegraphics[scale=0.20]{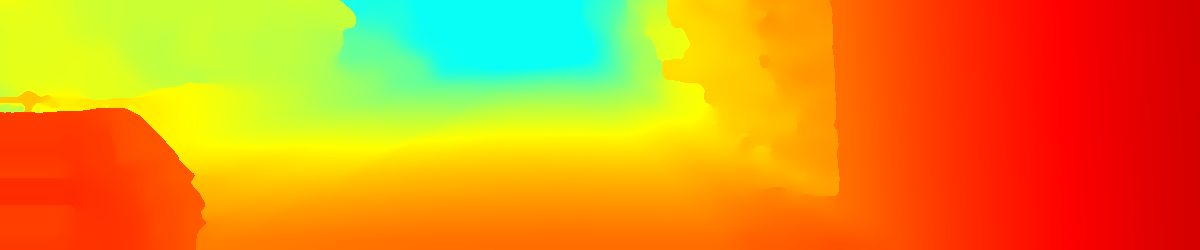} \\
\includegraphics[scale=0.20]{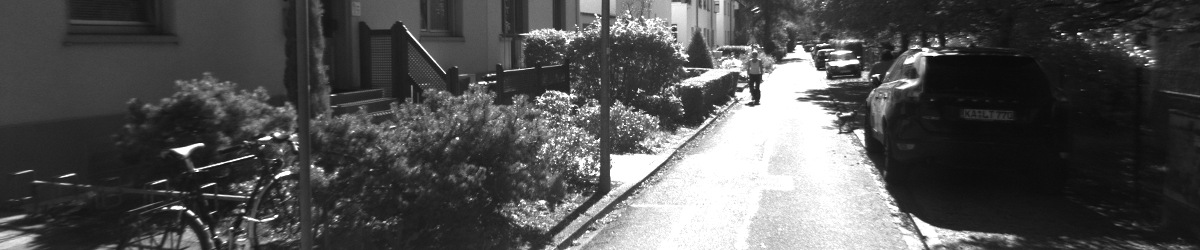} & \includegraphics[scale=0.20]{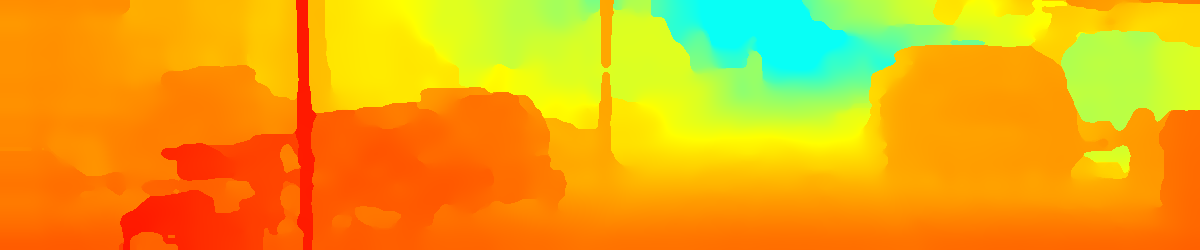} \\
\includegraphics[scale=0.20]{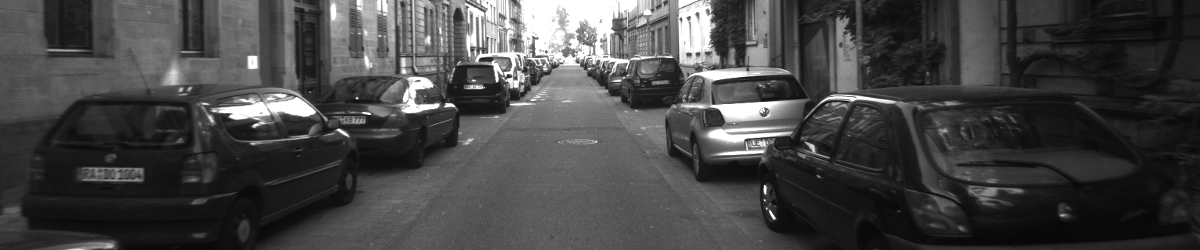} & \includegraphics[scale=0.20]{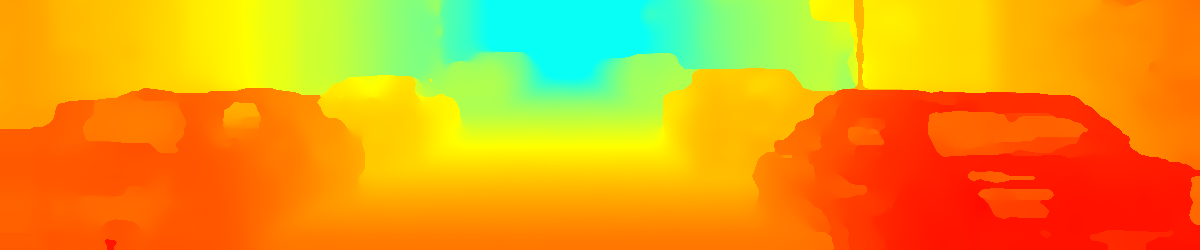} \\
\includegraphics[scale=0.20]{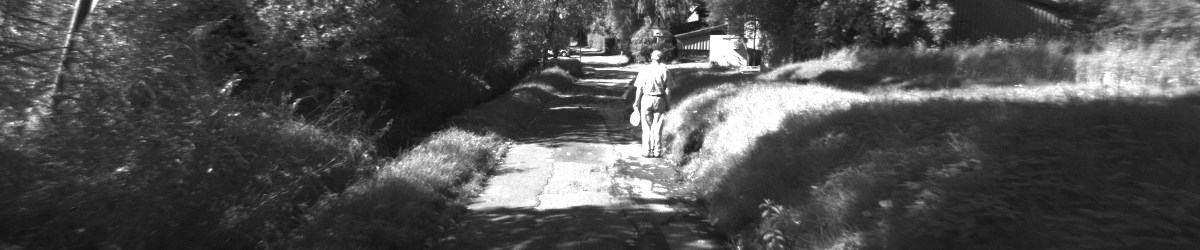} & \includegraphics[scale=0.20]{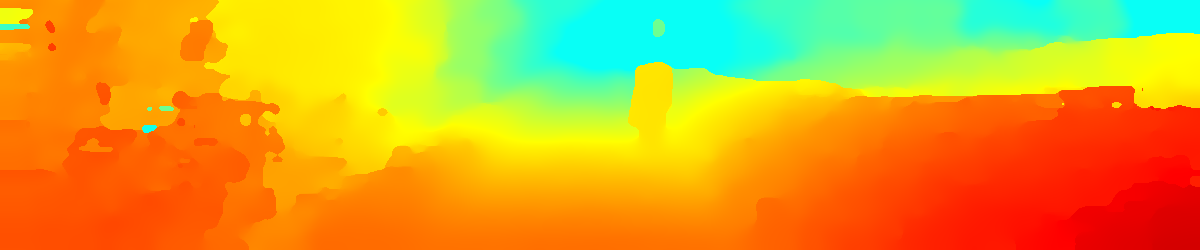} \\
\includegraphics[scale=0.20]{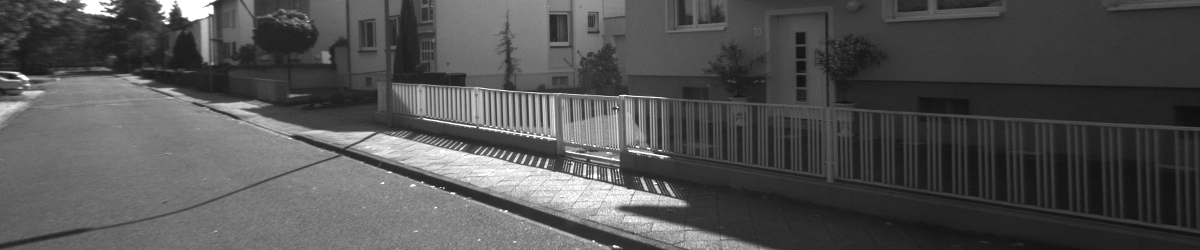} & \includegraphics[scale=0.20]{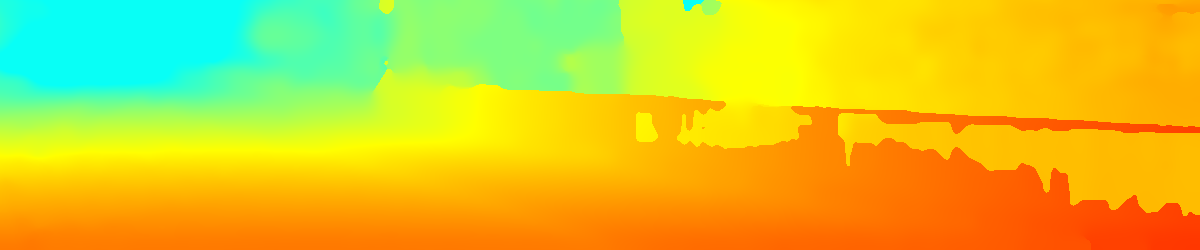} \\
\includegraphics[scale=0.20]{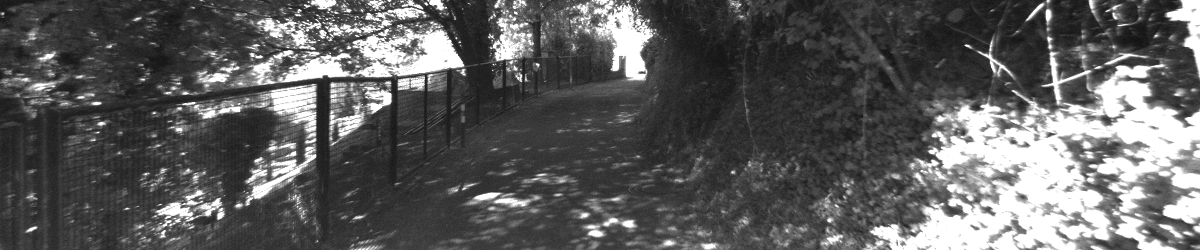} & \includegraphics[scale=0.20]{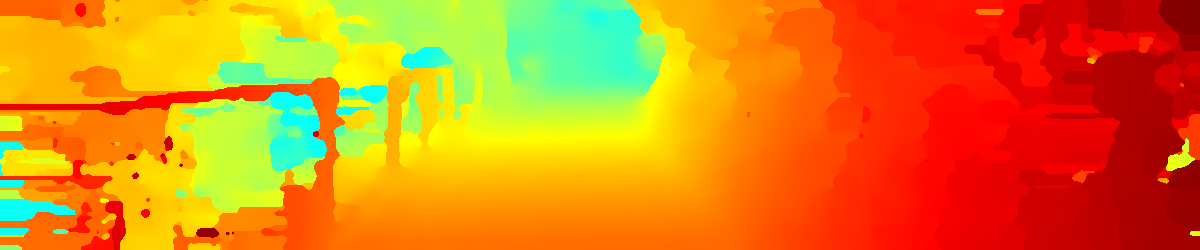} \\
\includegraphics[scale=0.20]{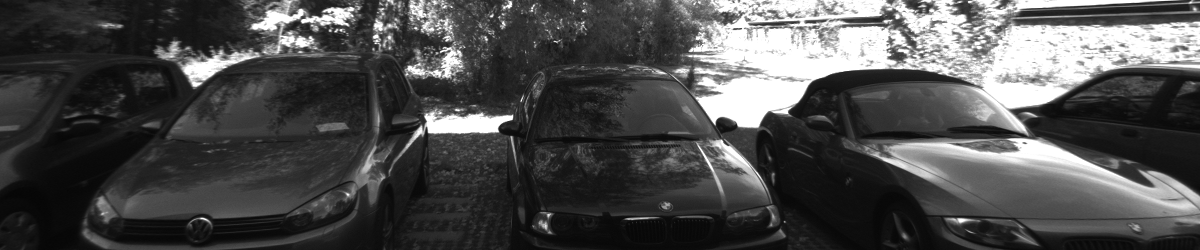} & \includegraphics[scale=0.20]{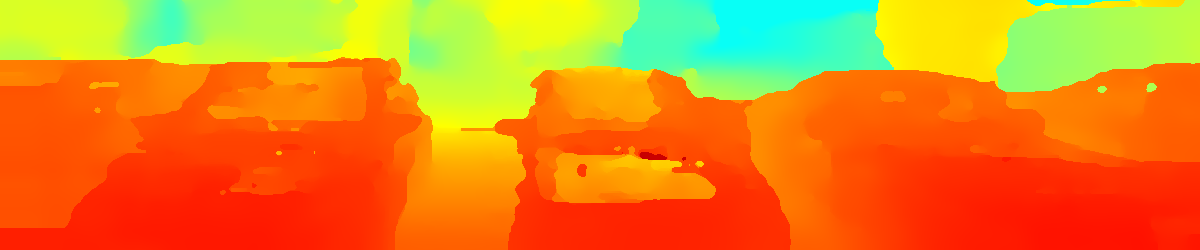} \\
\end{tabular}
\end{center}
\caption{The left column displays the left input image, while the right column
displays the output of our stereo method. Examples are sorted by difficulty,
with easy examples appearing at the top. Some of the difficulties include
reflective surfaces, occlusions, as well as regions with many jumps in
disparity, \eg fences and shrubbery. The examples towards the bottom were
selected to highlight the flaws in our method and to demonstrate the inherent
difficulties of stereo matching on real-world images.}

\label{fig:pred}
\end{figure*}

A selected set of examples, together with predictions from our method, are
shown in Figure~\ref{fig:pred}.

%

\subsection{Runtime}
We measure the runtime of our implementation on a computer with a Nvidia GeForce
GTX Titan GPU. Training takes 5 hours. Predicting a single image
pair takes 100 seconds. It is evident from Table~\ref{tab:kitti_runtime} that
the majority of time during prediction is spent in the forward pass of the
convolutional neural network.
\begin{table}[ht]
\begin{center}
\begin{tabular}{ll}
\textbf{Component} & \textbf{Runtime} \\\hline
Convolutional neural network & 95 s \\
Semiglobal matching & 3 s \\
Cross-based cost aggregation & 2 s \\
Everything else & 0.03 s
\end{tabular}
\end{center}
\caption{Time required for prediction of each component.}
\label{tab:kitti_runtime}
\end{table}

\subsection{Training set size}

\begin{figure}[ht]
\begin{center}
\includegraphics[scale=0.65]{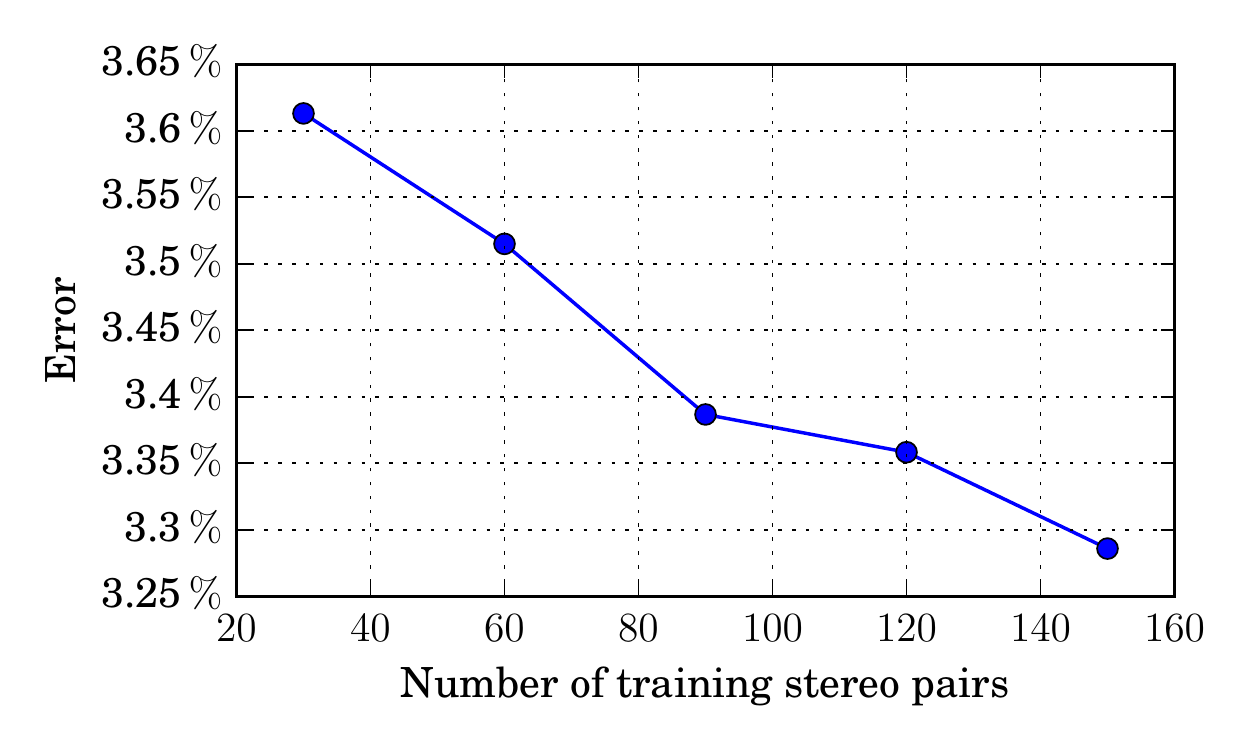}
\end{center}
\caption{The error on the test set as a function of the number of stereo pairs in the training set.}
\label{fig:tr_size}
\end{figure}

We would like to know if more training data would lead to a better stereo method.
To answer this question, we train our convolutional neural network on many
instances of the KITTI stereo dataset while varying the training set size.
The results of the experiment are depicted in Figure~\ref{fig:tr_size}.
We observe an almost linear relationship between the training set size and
error on the test set. These results imply that our method will improve as
larger datasets become available in the future.

\section{Conclusion}

Our result on the KITTI stereo dataset seems to suggest that convolutional
neural networks are a good fit for computing the stereo matching cost.
Training on bigger datasets will reduce the error rate even further.  Using
supervised learning in the stereo method itself could also be beneficial. Our
method is not yet suitable for real-time applications such as robot navigation.
Future work will focus on improving the network's runtime performance.

\bibliographystyle{apalike}
\bibliography{paper}

\begin{thebibliography}{}

\bibitem[Collobert et~al., 2011]{collobert2011torch7}
Collobert, R., Kavukcuoglu, K., and Farabet, C. (2011).
\newblock Torch7: A matlab-like environment for machine learning.
\newblock In {\em BigLearn, NIPS Workshop}, number EPFL-CONF-192376.

\bibitem[Geiger et~al., 2013]{Geiger2013IJRR}
Geiger, A., Lenz, P., Stiller, C., and Urtasun, R. (2013).
\newblock Vision meets robotics: The {KITTI} dataset.
\newblock {\em International Journal of Robotics Research (IJRR)}.

\bibitem[Haeusler et~al., 2013]{haeusler2013ensemble}
Haeusler, R., Nair, R., and Kondermann, D. (2013).
\newblock Ensemble learning for confidence measures in stereo vision.
\newblock In {\em Computer Vision and Pattern Recognition (CVPR), 2013 IEEE
  Conference on}, pages 305--312. IEEE.

\bibitem[Hirschmuller, 2008]{hirschmuller2008stereo}
Hirschmuller, H. (2008).
\newblock Stereo processing by semiglobal matching and mutual information.
\newblock {\em Pattern Analysis and Machine Intelligence, IEEE Transactions
  on}, 30(2):328--341.

\bibitem[Hirschmuller and Scharstein, 2009]{hirschmuller2009evaluation}
Hirschmuller, H. and Scharstein, D. (2009).
\newblock Evaluation of stereo matching costs on images with radiometric
  differences.
\newblock {\em Pattern Analysis and Machine Intelligence, IEEE Transactions
  on}, 31(9):1582--1599.

\bibitem[Kong and Tao, 2004]{kong2004method}
Kong, D. and Tao, H. (2004).
\newblock A method for learning matching errors for stereo computation.
\newblock In {\em BMVC}, pages 1--10.

\bibitem[Kong and Tao, 2006]{kong2006stereo}
Kong, D. and Tao, H. (2006).
\newblock Stereo matching via learning multiple experts behaviors.
\newblock In {\em BMVC}, pages 97--106.

\bibitem[Krizhevsky et~al., 2012]{krizhevsky2012imagenet}
Krizhevsky, A., Sutskever, I., and Hinton, G. (2012).
\newblock Imagenet classification with deep convolutional neural networks.
\newblock In {\em Advances in Neural Information Processing Systems 25}, pages
  1106--1114.

\bibitem[LeCun et~al., 1998]{lecun1998gradient}
LeCun, Y., Bottou, L., Bengio, Y., and Haffner, P. (1998).
\newblock Gradient-based learning applied to document recognition.
\newblock {\em Proceedings of the IEEE}, 86(11):2278--2324.

\bibitem[Li and Huttenlocher, 2008]{li2008learning}
Li, Y. and Huttenlocher, D.~P. (2008).
\newblock Learning for stereo vision using the structured support vector
  machine.
\newblock In {\em Computer Vision and Pattern Recognition, 2008. CVPR 2008.
  IEEE Conference on}, pages 1--8. IEEE.

\bibitem[Mei et~al., 2011]{mei2011building}
Mei, X., Sun, X., Zhou, M., Wang, H., Zhang, X., et~al. (2011).
\newblock On building an accurate stereo matching system on graphics hardware.
\newblock In {\em Computer Vision Workshops (ICCV Workshops), 2011 IEEE
  International Conference on}, pages 467--474. IEEE.

\bibitem[Peris et~al., 2012]{peris2012towards}
Peris, M., Maki, A., Martull, S., Ohkawa, Y., and Fukui, K. (2012).
\newblock Towards a simulation driven stereo vision system.
\newblock In {\em Pattern Recognition (ICPR), 2012 21st International
  Conference on}, pages 1038--1042. IEEE.

\bibitem[Scharstein and Pal, 2007]{scharstein2007learning}
Scharstein, D. and Pal, C. (2007).
\newblock Learning conditional random fields for stereo.
\newblock In {\em Computer Vision and Pattern Recognition, 2007. CVPR'07. IEEE
  Conference on}, pages 1--8. IEEE.

\bibitem[Scharstein and Szeliski, 2002]{scharstein2002taxonomy}
Scharstein, D. and Szeliski, R. (2002).
\newblock A taxonomy and evaluation of dense two-frame stereo correspondence
  algorithms.
\newblock {\em International journal of computer vision}, 47(1-3):7--42.

\bibitem[Spyropoulos et~al., 2014]{spyropoulos2014learning}
Spyropoulos, A., Komodakis, N., and Mordohai, P. (2014).
\newblock Learning to detect ground control points for improving the accuracy
  of stereo matching.
\newblock In {\em Computer Vision and Pattern Recognition (CVPR), 2014 IEEE
  Conference on}, pages 1621--1628. IEEE.

\bibitem[Vogel et~al., 2014]{vogel2014view}
Vogel, C., Roth, S., and Schindler, K. (2014).
\newblock View-consistent 3d scene flow estimation over multiple frames.
\newblock In {\em Computer Vision--ECCV 2014}, pages 263--278. Springer.

\bibitem[Vogel et~al., 2013]{vogel2013piecewise}
Vogel, C., Schindler, K., and Roth, S. (2013).
\newblock Piecewise rigid scene flow.
\newblock In {\em Computer Vision (ICCV), 2013 IEEE International Conference
  on}, pages 1377--1384. IEEE.

\bibitem[Yamaguchi et~al., 2012]{yamaguchi2012continuous}
Yamaguchi, K., Hazan, T., McAllester, D., and Urtasun, R. (2012).
\newblock Continuous markov random fields for robust stereo estimation.
\newblock In {\em Computer Vision--ECCV 2012}, pages 45--58. Springer.

\bibitem[Yamaguchi et~al., 2013]{yamaguchi2013robust}
Yamaguchi, K., McAllester, D., and Urtasun, R. (2013).
\newblock Robust monocular epipolar flow estimation.
\newblock In {\em Computer Vision and Pattern Recognition (CVPR), 2013 IEEE
  Conference on}, pages 1862--1869. IEEE.

\bibitem[Yamaguchi et~al., 2014]{yamaguchi2014efficient}
Yamaguchi, K., McAllester, D., and Urtasun, R. (2014).
\newblock Efficient joint segmentation, occlusion labeling, stereo and flow
  estimation.
\newblock In {\em Computer Vision--ECCV 2014}, pages 756--771. Springer.

\bibitem[Zhang et~al., 2009]{zhang2009cross}
Zhang, K., Lu, J., and Lafruit, G. (2009).
\newblock Cross-based local stereo matching using orthogonal integral images.
\newblock {\em Circuits and Systems for Video Technology, IEEE Transactions
  on}, 19(7):1073--1079.

\bibitem[Zhang and Seitz, 2007]{zhang2007estimating}
Zhang, L. and Seitz, S.~M. (2007).
\newblock Estimating optimal parameters for mrf stereo from a single image
  pair.
\newblock {\em Pattern Analysis and Machine Intelligence, IEEE Transactions
  on}, 29(2):331--342.

\end{thebibliography}

\end{document}